\documentclass[10pt,twocolumn,letterpaper]{article}

\usepackage{iccv}
\usepackage{times}
\usepackage{epsfig}
\usepackage{graphicx}
\usepackage{amsmath}
\usepackage{amssymb}
\usepackage{epstopdf}
\usepackage{amsthm}
\usepackage{dsfont}
\usepackage{multirow}
\usepackage{pifont}
\usepackage{color}
\usepackage{algorithmicx}
\usepackage{algorithm}
\usepackage{algpseudocode}
\newcommand{\cmark}{\ding{51}}%
\newcommand{\xmark}{\ding{55}}%

\iccvfinalcopy % *** Uncomment this line for the final submission

\def\iccvPaperID{2} % *** Enter the ICCV Paper ID here

\ificcvfinal\fi
\begin{document}

%%%%%%%%% TITLE
\title{Inferring Human Activities Using Robust Privileged Probabilistic Learning}

\author{Michalis Vrigkas$^1$ \qquad\hfill Evangelos Kazakos$^2$ \qquad\hfill Christophoros Nikou$^2$ \qquad\hfill Ioannis A. Kakadiaris$^1$\\
$^1$Computational Biomedicine Lab, University of Houston, Houston, TX, USA\\
$^2$Dept. Computer Science $\&$ Engineering, University of
Ioannina, Ioannina, Greece\\
}

\maketitle
\thispagestyle{empty}

%%%%%%%%% ABSTRACT
\begin{abstract}
Classification models may often suffer from ``structure
imbalance'' between training and testing data that may occur
due to the deficient data collection process. This imbalance
can be represented by the learning using privileged information
(LUPI) paradigm. In this paper, we present a supervised
probabilistic classification approach that integrates LUPI into
a hidden conditional random field (HCRF) model. The proposed
model is called LUPI-HCRF and is able to cope with additional
information that is only available during training. Moreover,
the proposed method employes Student's \textit{t}-distribution
to provide robustness to outliers by modeling the conditional
distribution of the privileged information. Experimental
results in three publicly available datasets demonstrate the
effectiveness of the proposed approach and improve the
state-of-the-art in the LUPI framework for recognizing human
activities.
\end{abstract}

%%%%%%%%% BODY TEXT
\section{Introduction}
\label{sec:intro}

The rapid development of human activity recognition systems for
applications such as surveillance and human-machine
interactions \cite{Cohen04,Smeulders13} brings forth the need
for developing new learning techniques. Learning using
privileged information (LUPI) \cite{Lopez_PazBSV16,sarafianos2016predicting,Vapnik09} has recently
generated considerable research interest. The insight of LUPI
is that one may have access to additional information about the
training samples, which is not available during testing. 

\begin{figure}[t]
\begin{center}
   \includegraphics[width=0.89\linewidth, clip=true]{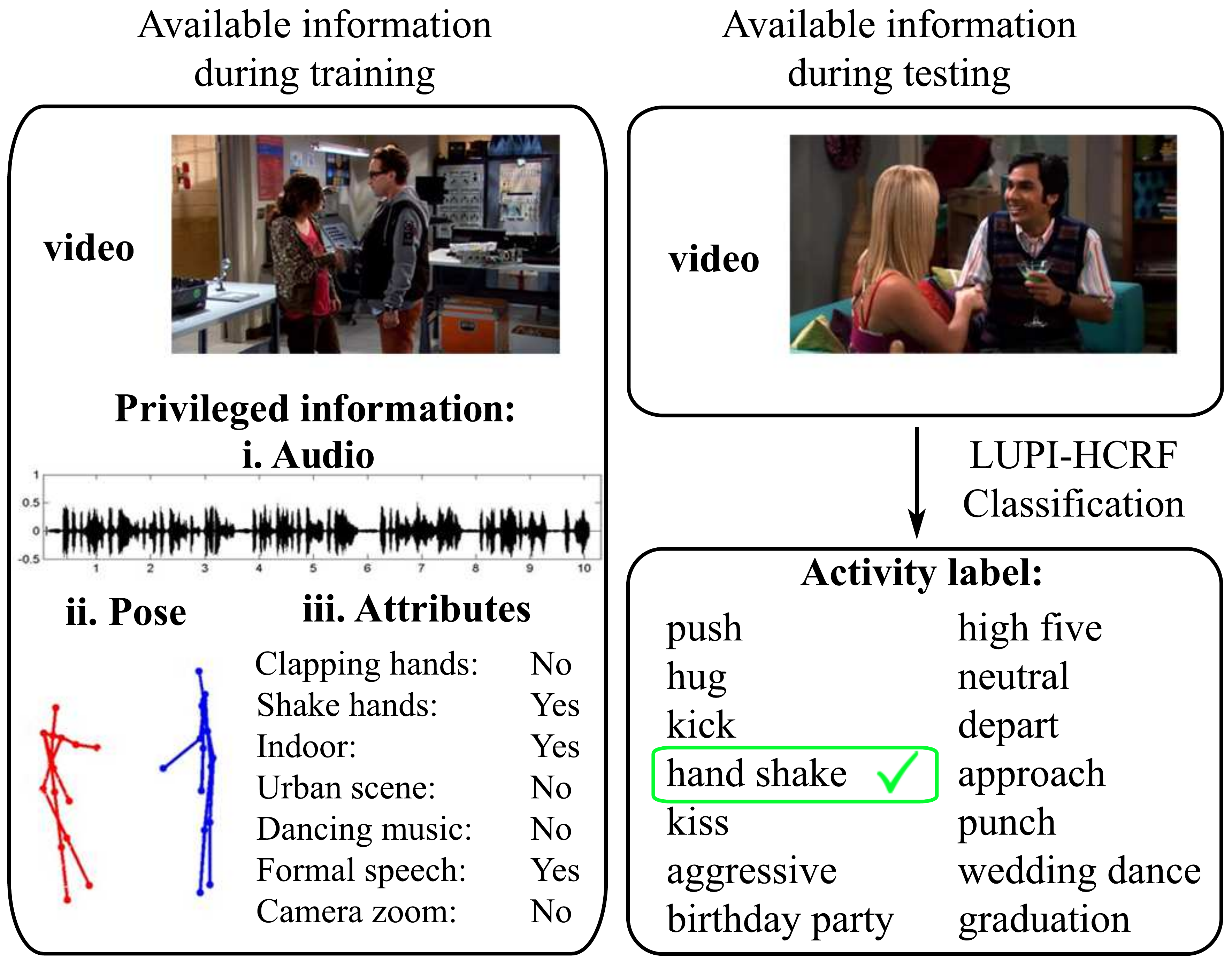}
\end{center}
   \caption{Robust learning using privileged information. Given a set of training
   examples and a set of additional information about the training samples
   (left) our system can successfully recognize the class label of the
   underlying activity without having access to the additional information
   during testing (right). We explore three different forms of privileged
   information (\eg, audio signals, human poses, and attributes) by
   modeling them with a Student's \textit{t}-distribution and incorporating
   them into the LUPI-HCRF model.}
\label{fig:figure1}
\end{figure}

Despite the impressive progress that has been made in
recognizing human activities, the problem still remains
challenging. First, constructing a visual model for learning
and analyzing human movements is difficult. The large
intra-class variabilities or changes in appearance make the
recognition problem difficult to address. Finally, the lack of
informative data or the presence of misleading information may
lead to ineffective approaches.

We address these issues by presenting a probabilistic approach,
which is able to learn human activities by exploiting
additional information about the input data, that may reflect
on auxiliary properties about classes and members of the
classes of the training data (Fig. \ref{fig:figure1}). In this
context, we employ a new learning method based on hidden
conditional random fields (HCRFs) \cite{Quattoni07}, called
LUPI-HCRF, which can efficiently manage dissimilarities in
input data, such as noise, or missing data, using a Student's
\textit{t}-distribution. The use of Student's
\textit{t}-distribution is justified by the property that it
has heavier tails than a standard Gaussian distribution, thus
providing robustness to outliers \cite{Peel00}.

The main contributions of our work can be summarized in the
following points. First, we developed a probabilistic human
activity recognition method that exploits privileged
information based on HCRFs to deal with missing or incomplete
data during testing. Second, contrary to previous methods,
which may be sensitive to outlying data measurements, we
propose a robust framework by employing a Student's
\textit{t}-distribution to attain robustness against outliers.
Finally, we emphasize the generic nature of our approach to
cope with samples from different modalities.

\section{Related work}
\label{sec:realatedwork}

A major family of methods relies on learning human activities
by building visual models and assigning activity roles to
people associated with an event \cite{Ramanathan13a,WangS13a}.
Earlier approaches use different kinds of modalities, such as
audio information, as additional information to construct
better classification models for activity recognition
\cite{SongMD12}.

A shared representation of human poses and visual information
has also been explored \cite{Ykiwon12}. Several kinematic
constraints for decomposing human poses into separate limbs
have been explored to localize the human body
\cite{CherianMAS14}. However, identifying which body parts are
most significant for recognizing complex human activities still
remains a challenging task \cite{LilloSN14}. Much focus has
also been given in recognizing human activities from movies or
TV shows by exploiting scene contexts to localize and
understand human interactions \cite{HoaiZ14,Perez12}. The
recognition accuracy of such complex videos can also be
improved by relating textual descriptions and visual context to
a unified framework \cite{RamanathanLL13}.

Recently, intermediate semantic features representation for
recognizing unseen actions during training has been proposed
\cite{JLiu11,Wang2010}. These features are learned during
training and enable parameter sharing between classes by
capturing the correlations between frequently occurring
low-level features \cite{AkataPHS13}. Instead of learning one
classifier per attribute, a two-step classification method has
been proposed by Lampert \etal \cite{Lampert09}. Specific
attributes are predicted from pre-trained classifiers and
mapped into a class-level score.

Recent methods that exploited deep neural networks have
demonstrated remarkable results in large-scale datasets.
Donahue \etal \cite{lrcn2014} proposed a recurrent
convolutional architecture, where recurrent long-term models
are connected to convolutional neural networks (CNNs) that can
be jointly trained to simultaneously learn spatio-temporal
dynamics. Wang \etal \cite{Wang_2015_CVPR} proposed a new video
representation that employs CNNs to learn multi-scale
convolutional feature maps. Tran \etal. \cite{tran2015learning}
introduced a 3D ConvNet architecture that learns
spatio-temporal features using 3D convolutions. A novel video
representation, that can summarize a video into a single image
by applying rank pooling on the raw image pixels, was proposed
by Bilen \etal \cite{Bilen_2016_CVPR}. Feichtenhofer \etal
\cite{Feichtenhofer_2016_CVPR} introduced a novel architecture
for two stream ConvNets and studied different ways for
spatio-temporal fusion of the ConvNet towers. Zhu \etal
\cite{Zhu_2016_CVPR} argued that videos contain one or more key
volumes that are discriminative and most volumes are irrelevant
to the recognition process.

The LUPI paradigm was first introduced by Vapnik and Vashist
\cite{Vapnik09} as a new classification setting to model based
on a max-margin framework, called SVM+. The choice of different
types of privileged information in the context of an object
classification task implemented in a max-margin scheme was also
discussed by Sharmanska \etal \cite{SharmanskaQL13}. Wand and
Ji \cite{ZWang15} proposed two different loss functions that
exploit privileged information and can be used with any
classifier. Recently, a combination of the LUPI framework and
active learning has been explored by Vrigkas \etal
\cite{MVrigkas_ICIP16} to classify human activities in a
semi-supervised scheme.

%------------------------------------------------------------------------
\section{Robust privileged probabilistic learning}
\label{sec:approach}

\begin{figure}[t]
\begin{center}
    \includegraphics[width=0.83\linewidth, clip=true]{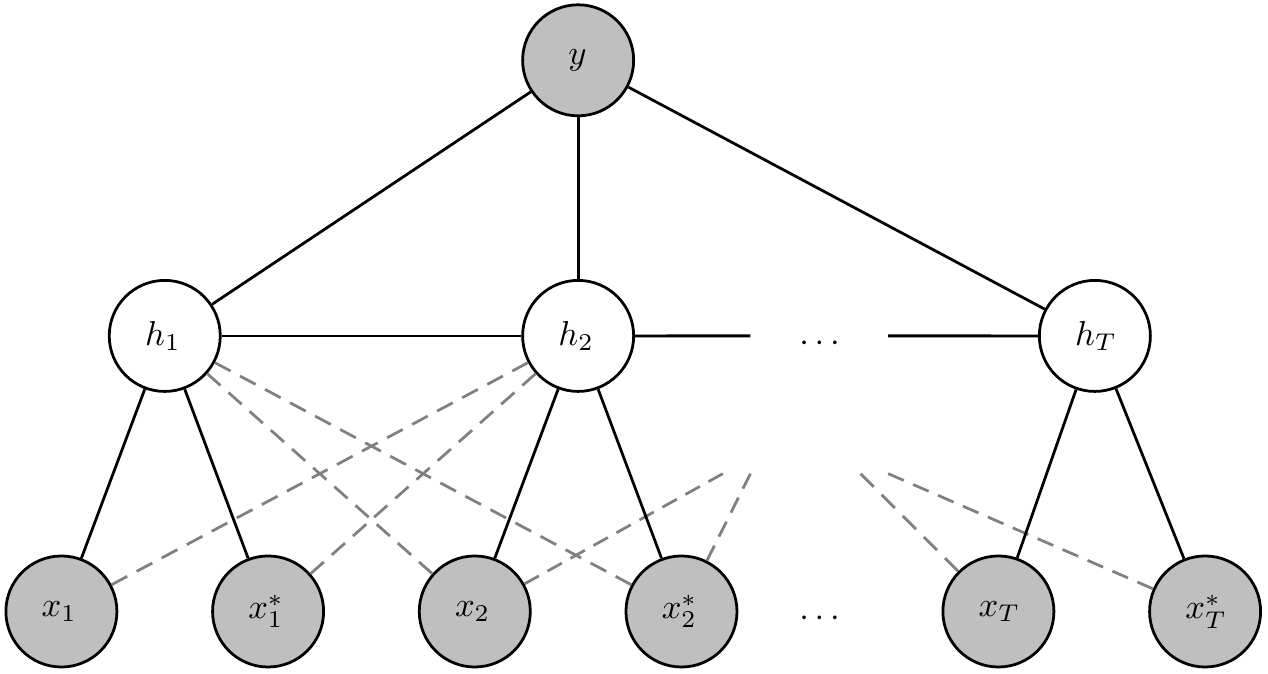}
\end{center}
    \caption{Graphical representation of the chain structure model.
    The grey nodes are the observed features ($x_{i}$), the privileged
    information ($x^{*}_{i}$), and the unknown labels ($y$), respectively.
    The white nodes are the unobserved hidden variables ($h$).}
\label{fig:figure2}
\end{figure}

Our method uses HCRFs, which are defined by a chained
structured undirected graph $\mathcal{G} = (\mathcal{V},
\mathcal{E})$ (see Fig. \ref{fig:figure2}), as the
probabilistic framework for modeling the activity of a subject
in a video. During training, a classifier and the mapping from
observations to the label set are learned. In testing, a probe
sequence is classified into its respective state using loopy
belief propagation (LBP) \cite{Komodakis07}.

%%%----------------------------------------------------------------
%%%----------------------------------------------------------------

%%%~~~~~~~~~~~~~~~~~~~~~~~~~~~~~~~~~~~~~~~~~~~~~~~~~~~~~~~~~~~~~~~~~~~~%%%
\subsection{LUPI-HCRF model formulation}
\label{subsec:model}
%%%~~~~~~~~~~~~~~~~~~~~~~~~~~~~~~~~~~~~~~~~~~~~~~~~~~~~~~~~~~~~~~~~~~~~%%%

We consider a labeled dataset with $N$ video sequences
consisting of triplets $\mathcal{D} =\{(\mathbf{x}_{i,j},
\mathbf{x}^{*}_{i,j}, y_{i})\}_{i=1}^{N}$, where
$\mathbf{x}_{i,j} \in \mathbb{R}^{M_{\mathbf{x}} \times T}$ is
an observation sequence of length $T$ with $j=1 \ldots T$. For
example, $\mathbf{x}_{i,j}$ might correspond to the
$j^\text{th}$ frame of the $i^\text{th}$  video sequence.
Furthermore, $y_{i}$ corresponds to a class label defined in a
finite label set $\mathcal{Y}$. Also, the additional
information about the observations $\mathbf{x}_{i}$ is encoded
in a feature vector $\mathbf{x}^{*}_{i,j} \in
\mathbb{R}^{M_{\mathbf{x}^{*}} \times T}$. Such privileged
information is provided only at the training step and  it is
not available during testing. Note that we do not make any
assumption about the form of the privileged data. In what
follows, we omit indices $i$ and $j$ for simplicity.

The LUPI-HCRF model is a member of the exponential family and
the probability of the class label given an observation
sequence is given by:
\begin{equation}
\label{eq:model.1}
\begin{split}
p(y|\mathbf{x},\mathbf{x}^{*};\mathbf{w}) &=
\displaystyle \sum_{\mathbf{h}} p(y,\mathbf{h}|\mathbf{x},\mathbf{x}^{*};\mathbf{w}) \\
&=\sum_{\mathbf{h}}\exp\left(E(y,\mathbf{h}|\mathbf{x},\mathbf{x}^{*};\mathbf{w}) -
A(\mathbf{w})\right)\, , \raisetag{2.5\baselineskip}
\end{split}
\end{equation}
where $\mathbf{w} = [\boldsymbol\theta, \boldsymbol\omega]$ is
a vector of model parameters, and \mbox{$\mathbf{h} =
\{h_{1},h_{2},\ldots,h_{T}\}$}, with $h_{j} \in \mathcal{H}$
being a set of latent variables. In particular, the number of
latent variables may be different from the number of samples,
as $h_{j}$ may correspond to a substructure in an observation.
Moreover, the features follow the structure of the graph, in
which no feature may depend on more than two hidden states
$h_{j}$ and $h_{k}$ \cite{Quattoni07}. This property not only
captures the synchronization points between the different sets
of information of the same state, but also models the
compatibility between pairs of consecutive states. We assume
that our model follows the first-order Markov chain structure
(\ie., the current state affects the next state). Finally,
$E(y,\mathbf{h}|\mathbf{x};\mathbf{w})$ is a vector of
sufficient statistics and $A(\mathbf{w})$ is the log-partition
function ensuring normalization:
\begin{equation}
\label{eq:model.2}
A(\mathbf{w}) = \log \sum_{y'} \sum_{\mathbf{h}}
\exp\left(E(y',\mathbf{h}|\mathbf{x},\mathbf{x}^{*};\mathbf{w})\right)\, .
\end{equation}

Different sufficient statistics
$E(y|\mathbf{x},\mathbf{x}^{*};\mathbf{w})$ in
\eqref{eq:model.1} define different distributions. In the
general case, sufficient statistics consist of indicator
functions for each possible configuration of unary and pairwise
terms:
\begin{equation}
\label{eq:model.3}
\begin{split}
\resizebox{1.0\hsize}{!}{$\displaystyle E(y,\mathbf{h}|\mathbf{x},\mathbf{x}^{*};\mathbf{w})
 = \sum_{j \in \mathcal{V}}  \Phi(y,h_{j},\mathbf{x}_{j},\mathbf{x}_{j}^{*};\boldsymbol\theta)
 + \sum_{j, k \in \mathcal{E}} \Psi(y,h_{j},h_{k};\boldsymbol\omega)\, ,$}
\end{split}
\end{equation}
where the parameters $\boldsymbol\theta$ and
$\boldsymbol\omega$ are the unary and the pairwise weights,
respectively, that need to be learned. The unary potential does
not depend on more than two hidden variables $h_{j}$ and
$h_{k}$, and the pairwise potential may depend on $h_{j}$ and
$h_{k}$, which means that there must be an edge $(j,k)$ in the
graphical model. The unary potential is expressed by:
\begin{equation}
\label{eq:model.3a}
\begin{split}
\Phi(y,h_{j},\mathbf{x}_{j},&\mathbf{x}_{j}^{*};\boldsymbol\theta) =
\sum\limits_{\ell} \phi_{1,\ell}(y,h_{j};\boldsymbol\theta_{1,\ell})\\
& + \phi_{2}(h_{j},\mathbf{x}_{j};\boldsymbol\theta_{2})
+ \phi_{3}(h_{j},\mathbf{x}_{j}^{*};\boldsymbol\theta_{3})\, ,
\raisetag{2.5\baselineskip}
\end{split}
\end{equation}
and it can be seen as a state function, which consists of three
different feature functions. The label feature function, which models
the relationship between the label $y$ and the hidden variables
$h_{j}$, is expressed by:
\begin{equation}
\label{eq:model.3b}
\phi_{1,\ell}(y,h_{j};\boldsymbol\theta_{1,\ell})
= \sum\limits_{\lambda \in \mathcal{Y}}
\sum\limits_{a \in \mathcal{H}}
\boldsymbol\theta_{1,\ell} \mathds{1}(y =
\lambda) \mathds{1}(h_{j} = a)\, ,
\end{equation}
where $\mathds{1}(\cdot)$ is the indicator function, which is equal
to $1$ if its argument is true, and $0$ otherwise. 
The observation feature function, which models the relationship between
the hidden variables $h_{j}$ and the observations $\mathbf{x}$, is
defined by:
\begin{equation}
\label{eq:model.3c}
\phi_{2}(h_{j},\mathbf{x}_{j};\boldsymbol\theta_{2})
= \sum\limits_{a \in \mathcal{H}}
\boldsymbol\theta_{2}^{\top}\mathds{1}(h_{j} = a)
\mathbf{x}_{j}\, .
\end{equation}
Finally, the
privileged feature function, which models the relationship between
the hidden variables $h_{j}$ is defined by: 
\begin{equation}
\label{eq:model.3d}
\phi_{3}(h_{j},\mathbf{x}_{j}^{*};\boldsymbol\theta_{3}) = \sum\limits_{a \in
\mathcal{H}} \boldsymbol\theta_{3}^{\top}\mathds{1}(h_{j} = a) \mathbf{x}_{j}^{*}\, .
\end{equation}

The pairwise potential is a transition function and represents the
association between a pair of connected hidden states $h_{j}$ and
$h_{k}$ and the label $y$. It is expressed by:
\begin{equation}
\label{eq:model.3e}
\resizebox{1.0\hsize}{!}{$\displaystyle \Psi(y,h_{j},h_{k};\boldsymbol\omega) = \sum\limits_{\substack{\lambda \in
\mathcal{Y} \\ a,b \in \mathcal{H}}} \sum_{\ell} \boldsymbol\omega_{\ell}
\mathds{1}(y =\lambda) \mathds{1}(h_{j} = a)\mathds{1}(h_{k} = b)\, .$}
\end{equation}

%%%~~~~~~~~~~~~~~~~~~~~~~~~~~~~~~~~~~~~~~~~~~~~~~~~~~~~~~~~~~~~~~~~~~~~%%%
\subsection{Parameter learning and inference}
\label{subsec:parameter}
%%%~~~~~~~~~~~~~~~~~~~~~~~~~~~~~~~~~~~~~~~~~~~~~~~~~~~~~~~~~~~~~~~~~~~~%%%
In the training step the optimal parameters $\mathbf{w}^{*}$
are estimated by maximizing the following loss function:
\begin{equation}
\label{eq:model.4}
L(\mathbf{w}) = \sum_{i=1}^{N} \log p(y_{i}|\mathbf{x}_{i},\mathbf{x}^{*}_{i};\mathbf{w})
- \frac{1}{2\sigma^{2}}\| \mathbf{w} \|^{2}\, .
\end{equation}

The first term is the log-likelihood of the posterior probability
$p(y|\mathbf{x},\mathbf{x}^{*};\mathbf{w})$ and quantifies how well
the distribution in Eq. \eqref{eq:model.1} defined by the parameter
vector $\mathbf{w}$ matches the labels $y$. 
The second term is a Gaussian prior with variance $\sigma^{2}$
and works as a regularizer. The loss function is optimized
using the limited-memory BFGS (LBFGS) method \cite{Nocedal06} to
minimize the negative log-likelihood of the data.

Our goal is to estimate the optimal label configuration over
the testing input, where the optimality is expressed in terms
of a cost function. To this end, we maximize the posterior
probability and marginalize over the latent variables
$\mathbf{h}$ and the privileged information $\mathbf{x}^{*}$:
\begin{equation}
\label{eq:model.6}
\begin{split}
\displaystyle y &= \operatorname*{arg\,max}_{y}
p(y|\mathbf{x};\mathbf{w})\\
&= \operatorname*{arg\,max}_{y}\sum_{\mathbf{h}}\sum_{\mathbf{x}^{*}}
p(y,\mathbf{h}|\mathbf{x},\mathbf{x}^{*};\mathbf{w})
p(\mathbf{x}^{*}|\mathbf{x};\mathbf{w})\, .
\end{split}
\end{equation}

To efficiently cope with outlying measurements about the
training data, we consider that the training samples
$\mathbf{x}$ and $\mathbf{x}^{*}$ jointly follow a Student's
\textit{t}-distribution. Therefore, the conditional
distribution $p(\mathbf{x}^{*}|\mathbf{x};\mathbf{w})$ is also
a Student's \textit{t}-distribution
$\text{St}(\mathbf{x}^{*}|\mathbf{x};\mu^{*},\Sigma^{*},\nu^{*})$,
where $\mathbf{x}^{*}$ forms the first $M_{{\mathbf{x}}^{*}}$
components of $\left( \mathbf{x}^{*}, \mathbf{x}\right)^{T}$,
$\mathbf{x}$ comprises the remaining $M - M_{\mathbf{x}^{*}}$
components, with mean vector $\mu^{*}$, covariance matrix
$\Sigma^{*}$ and $\nu^{*} \in [0, \infty)$ corresponds to the
degrees of freedom of the distribution \cite{Kotz04}. If the
data contain outliers, the degrees of freedom parameter
$\nu^{*}$ is weak and the mean and covariance of the data are
appropriately weighted in order not to take into account the
outliers. An approximate inference is employed for estimation
of the marginal probability (Eq. \eqref{eq:model.6}) by
applying the LBP algorithm \cite{Komodakis07}.

%------------------------------------------------------------------------
\section{Multimodal Feature Fusion}
\label{sec:mapping}

One drawback of combining features of different modalities is
the different frame rate that each modality may have. Thus,
instead of directly combining multimodal features together one
may employ canonical correlation analysis (CCA)
\cite{Hardoon04} to exploit the correlation between the
different modalities by projecting them onto a common subspace
so that the correlation between the input vectors is maximized
in the projected space. In this paper, we followed a different
approach. Our model is able to learn the relationship between
the input data and the privileged features. 
To this end, we jointly calibrate the different modalities by
learning a multiple output linear regression model
\cite{PalatucciPHM09}. Let $\mathbf{x} \in \mathbb{R}^{M \times
d}$ be the input raw data and $\mathbf{a} \in \mathbb{R}^{M
\times p}$ be the set of semantic attributes (privileged
features). Our goal is to find a set of weights
$\boldsymbol\gamma \in \mathbb{R}^{d \times p}$, which relates
the privileged features to the regular features by minimizing a
distance function across the input samples and their
attributes:
\begin{equation}
\label{eq:exper.1}
\operatorname*{arg\,min}_{\boldsymbol\gamma}\| \mathbf{x}\boldsymbol\gamma
- \mathbf{a} \|^{2} + \eta\| \boldsymbol\gamma \|^{2} \, ,
\end{equation}
where $\| \boldsymbol\gamma \|^{2}$ is a regularization term
and $\eta$ controls the degree of the regularization, which was
chosen to give the best solution by using cross validation with
$\eta \in [10^{-4},1]$. Following a constrained least squares
(CLS) optimization problem and minimizing $\| \boldsymbol\gamma
\|^{2}$ subject to $\mathbf{x}\boldsymbol\gamma = \mathbf{a}$,
Eq. \eqref{eq:exper.1} has a closed form solution
$\boldsymbol\gamma = \left(\mathbf{x}^{T}\mathbf{x} + \eta
I\right)^{-1} \mathbf{x}^{T}\mathbf{a}$, where $I$ is the
identity matrix. Note that the minimization of Eq.
\eqref{eq:exper.1} is fast since it needs to be solved only
once during training. Finally, we obtain the prediction $f$ of
the privileged features by multiplying the regular features
with the learned weights $f = \mathbf{x} \cdot
\boldsymbol\gamma$. The main steps of the proposed method are
summarized in Algorithm \ref{alg:hcrfplusalgo.1}.
%%%----------------------------------------------------------------
\begin{algorithm}[!t]
\caption{Robust privileged probabilistic leaning}
\label{alg:hcrfplusalgo.1}

\textbf{Input:~} Original data $\mathbf{x}$, privileged data
$\mathbf{x^{*}}$, class labels $y$
\begin{algorithmic}[1]
\State Perform feature extraction from both $\mathbf{x}$ and
$\mathbf{x^{*}}$

\State Project $\mathbf{x}$ and
$\mathbf{x^{*}}$ onto a common space using Eq. \eqref{eq:exper.1}

\State $\mathbf{w^{*}} \leftarrow
\displaystyle\operatorname*{arg\,min}_{\mathbf{w}}
\left(-L(\mathbf{w})\right)$ \,/* \textit{Train LUPI-HCRF on
$\mathbf{x}$ and $\mathbf{x^{*}}$ using Eq. \eqref{eq:model.4}} */

\State $\displaystyle \hat{y} \leftarrow
\operatorname*{arg\,max}_{y} p(y|\mathbf{x};\mathbf{w})$
\,/* \textit{Marginalize over $\mathbf{h}$ and $\mathbf{x^{*}}$
using Eq. \eqref{eq:model.6}} */
\end{algorithmic}
\textbf{Output:~} Predicted labels $\hat{y}$
\end{algorithm}
%%%----------------------------------------------------------------

%------------------------------------------------------------------------

\section{Experiments}
\label{sec:experiments}

\textbf{Datasets:} The TV human interaction (TVHI)
\cite{Perez12} dataset consists of $300$ videos and contains
four kinds of interactions. The SBU Kinect Interaction (SBU)
\cite{Ykiwon12} dataset is a collection of approximately $300$
videos that contain eight different interaction classes.
Finally, the unstructured social activity attribute (USAA)
\cite{FuHXG12} dataset includes eight different semantic class
videos of social occasions and contains around $100$ videos per
class for training and testing.

%%%~~~~~~~~~~~~~~~~~~~~~~~~~~~~~~~~~~~~~~~~~~~~~~~~~~~~~~~~~~~~~~~~~~~~%%%
\subsection{Implementation details}
\label{subsec:impDet}
%%%~~~~~~~~~~~~~~~~~~~~~~~~~~~~~~~~~~~~~~~~~~~~~~~~~~~~~~~~~~~~~~~~~~~~%%%
\textbf{Feature selection:} For the evaluation of our method,
we used spatio-temporal interest points (STIP) \cite{Laptev05}
as our base video representation. These features were selected
because they can capture salient visual motion patterns in an
efficient and compact way. In addition, for the TVHI dataset,
we also used the provided annotations, which are related to the
locations of the persons in each video clip. For this dataset,
we used audio features as privileged information. More
specifically, we employed the mel-frequency cepstral
coefficients (MFCC) \cite{Rabiner93} features, resulting in a
collection of $13$ MFCC coefficients, and their first and
second order derivatives forming a $39$-dimensional feature
vector.

Furthermore, for the SBU dataset, as privileged information, we
used the positions of the locations of the joints for each
person in each frame, and six more feature types concerning
joint distance, joint motion, plane, normal plane, velocity,
and normal velocity as described by Yun \etal \cite{Ykiwon12}.
Finally, for the USAA dataset we used the provided attribute
annotation as privileged information to characterize each class
with a feature vector of semantic attributes. As a
representation of the video data, we used the provided SIFT
\cite{Lowe04}, STIP, and MFCC features. Table \ref{Tab:table1}
summarizes all forms of features used either as regular or
privileged for each dataset during training and testing.

\begin{table}[!t]
\begin{center}
\resizebox{0.97\columnwidth}{!}{
\begin{tabular}{l|lcc}
\hline
Dataset & Features (dimension) & Regular & Privileged \\
\hline \hline
\multirow{3}{*}{TVHI \cite{Perez12}} & STIP ($162$) & \cmark & ~  \\
~ & Head orientations (2) & \cmark & ~ \\
~ & MFCC ($39$) & ~ & \cmark\\
\hline
\multirow{2}{*}{SBU \cite{Ykiwon12}} & STIP ($162$) & \cmark & ~ \\
~ & Pose ($15$) & ~ & \cmark \\
\hline
\multirow{4}{*}{USAA \cite{FuHXG12}} & STIP ($162$) & \cmark & ~ \\
~ & SIFT ($128$) & \cmark & ~ \\
~ & MFCC ($39$) & \cmark & ~\\
~ & Semantic attributes ($69$) & ~ & \cmark \\
\hline
\end{tabular}
}
\end{center}
\caption{Types of features used for human activity recognition for
each dataset. The numbers in parentheses indicate the dimension of
the features. The checkmark corresponds to the usage of the specific
information as regular or privileged. Privileged features are used
only during training.} \label{Tab:table1}
\end{table}

\textbf{Model selection:} The  model in Fig. \ref{fig:figure2}
was trained by varying the number of hidden states from $4$ to
$20$, with a maximum of $400$ iterations for the termination of
LBFGS. The $L_{2}$ regularization scale term $\sigma$ was
searched within $\{10^{-3}, \dots, 10^{3}\}$ and $5$-fold cross
validation was used to split the datasets into training and
test sets, and the average results over all the examined
configurations are reported.

%%%~~~~~~~~~~~~~~~~~~~~~~~~~~~~~~~~~~~~~~~~~~~~~~~~~~~~~~~~~~~~~~~~~~~~%%%
\subsection{Results and discussion}
\label{subsec:results}
%%%~~~~~~~~~~~~~~~~~~~~~~~~~~~~~~~~~~~~~~~~~~~~~~~~~~~~~~~~~~~~~~~~~~~~%%%

\textbf{Classification with hand-crafted features:} We compare the
results of our LUPI-HCRF method with the state-of-the-art SVM+
method \cite{Vapnik09} and other baselines that incorporate the
LUPI paradigm. Also, to demonstrate the efficacy of robust privileged
information to the problem of human activity recognition,
we compared it with ordinary SVM and HCRF, as if they could
access both the original and the privileged information at test
time. This means that we do not differentiate between regular
and privileged information, but use both forms of information
as regular to infer the underlying class label instead. Also,
for the SVM+ and SVM, we consider a one-versus-all
decomposition of multi-class classification scheme and average
the results for every possible configurations. Finally, the
optimal parameters for the SVM and SVM+ were selected using
cross validation.

\begin{figure}[!t]
\begin{center}
\begin{tabular}{cc}
    \multicolumn{2}{c}{\includegraphics[width=0.4\columnwidth, clip=true]{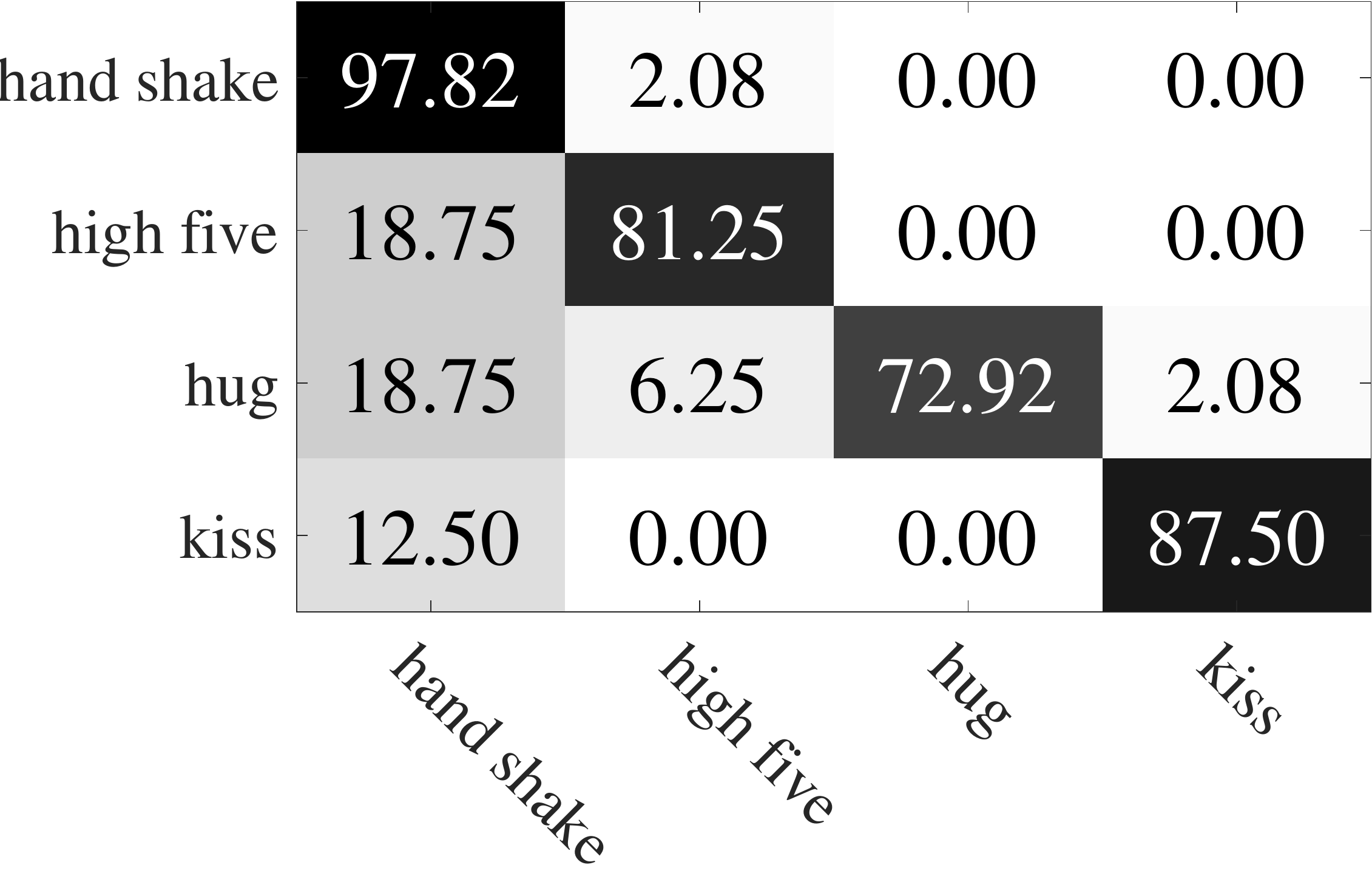} }\\
    \multicolumn{2}{c}{(a) } \\  
    \includegraphics[width=0.47\columnwidth, clip=true]{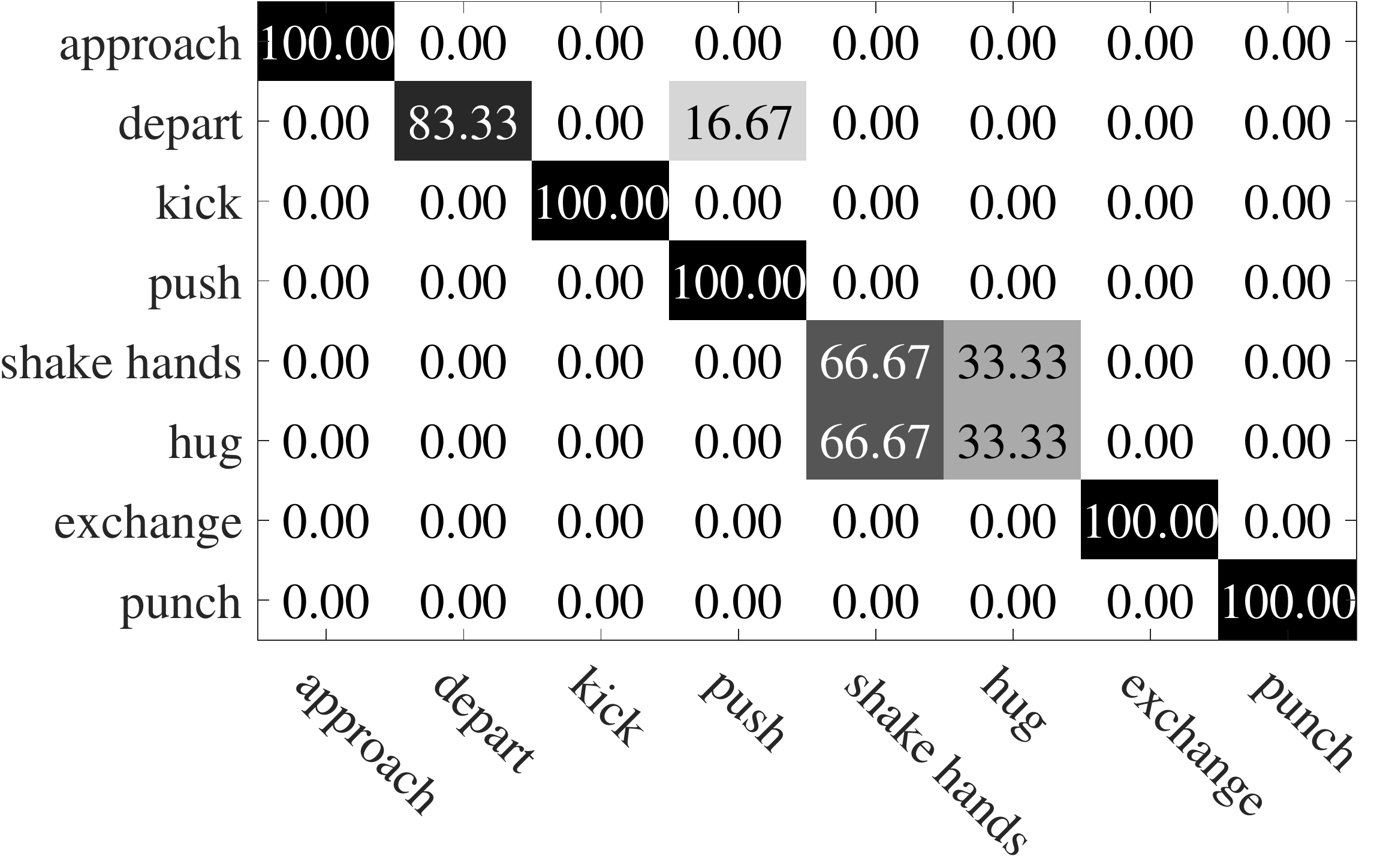} &
    \includegraphics[width=0.47\columnwidth, clip=true]{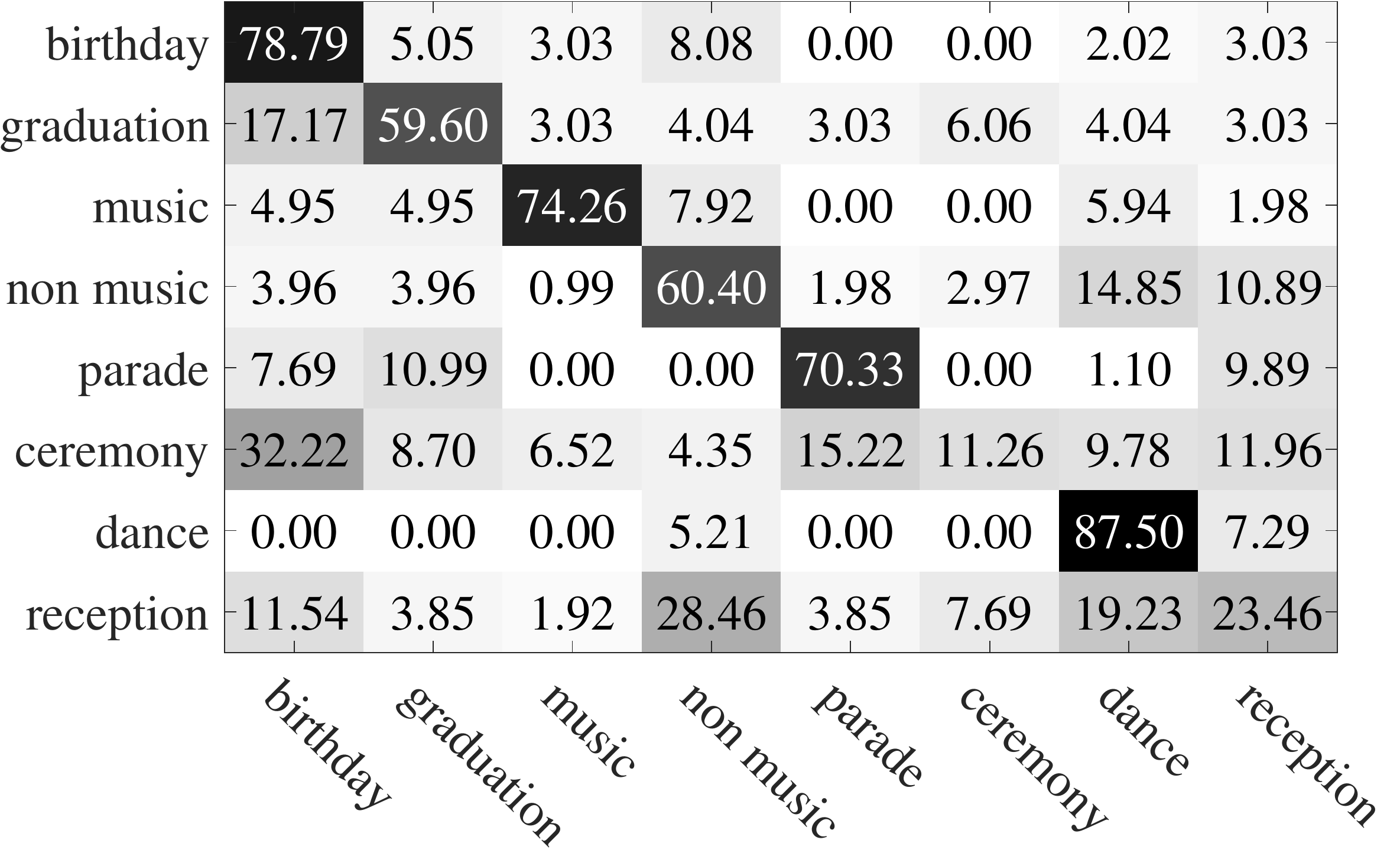} \\
    (b) & (c)
\end{tabular}
\end{center}
    \caption{Confusion matrices for the classification results of the
    proposed LUPI-HCRF approach for (a) the TVHI \cite{Perez12}, (b) the SBU
    \cite{Ykiwon12}, and (c) the USAA \cite{FuHXG12} datasets.}
\label{fig:figure4}
\end{figure}
The resulting confusion matrices for all datasets are depicted
in Fig. \ref{fig:figure4}. For the TVHI and SBU datasets, the
classification errors between different classes are relatively
small, as only a few classes are strongly confused with each
other. For the USAA dataset, the different classes may be
strongly confused (\eg, the class \emph{wedding ceremony} is
confused with the class \emph{graduation party}) as the dataset
has large intra-class variabilities, while the different
classes may share the same attribute representation as
different videos may have been captured under similar
conditions.

\begin{table}[!t]
\begin{center}
\resizebox{0.97\columnwidth}{!}{
\begin{tabular}{llcc}
\hline
Dataset & Regular & Privileged & Accuracy ($\%$) \\
\hline \hline
\multirow{4}{*}{TVHI \cite{Perez12}} & visual & \xmark & $60.9$ \\
~ & audio & \xmark & $35.9$ \\
~ & visual+audio & \xmark & $81.3$ \\
~ & visual & audio & $\mathbf{84.9}$ \\
\hline
\multirow{4}{*}{SBU \cite{Ykiwon12}} & visual & \xmark & $69.8$ \\
~ & pose & \xmark & $62.5$ \\
~ & visual+pose & \xmark & $81.4$ \\
~ & visual & pose & $\mathbf{85.4}$ \\
\hline
\multirow{4}{*}{USAA \cite{FuHXG12}} & visual & \xmark & $55.5$ \\
~ & sem. attributes & \xmark & $37.4$ \\
~ & visual+sem. attributes & \xmark & $54.0$ \\
~ & visual & sem. attributes & $\mathbf{58.1}$ \\
\hline
\end{tabular}
}
\end{center}
\caption{Comparison of feature combinations for classifying
human activities and events on TVHI \cite{Perez12}, SBU
\cite{Ykiwon12} and USAA \cite{FuHXG12} datasets. Robust
privileged information seems to work in favor of the
classification task. The crossmark corresponds to the absence
of privileged information during training.} \label{Tab:table2}
\end{table}
The benefit of using robust privileged information along with
conventional data instead of using each modality separately or
both modalities as regular information is shown in Table
\ref{Tab:table2}. The combination of regular and privileged
features has considerably increased the recognition accuracy to
much higher levels than using each modality separately. If only
privileged information is used as regular for classification,
the recognition accuracy is notably lower than when using
visual information for the classification task. Thus, we may
come to the conclusion that finding proper privileged
information is not always a straightforward problem.

\begin{table}[!t]
\begin{center}
\resizebox{0.99\columnwidth}{!}{
\begin{tabular}{l|ccc}
\hline
\multicolumn{4}{c}{Hand-crafted features}\\
\hline
Method & TVHI & SBU & USAA \\
\hline  \hline
Wang and Schmid \cite{WangS13a} & $76.1 \pm 0.4$ &  $79.6 \pm $0.4 & $55.5 \pm 0.1$ \\
Wang and Ji \cite{ZWang15} & $74.8 \pm 0.2$ & $62.4 \pm 0.3$ & $48.5 \pm 0.2$ \\
Sharmanska \etal \cite{SharmanskaQL13} & $65.2 \pm 0.1$ & $56.3 \pm 0.2$ & $56.3 \pm 0.2$ \\
SVM \cite{Bishop06} & $75.9 \pm 0.6$ &  $79.4 \pm 0.4$ & $47.4 \pm 0.1$ \\
HCRF \cite{Quattoni07} & $81.3 \pm 0.7$ & $81.4 \pm 0.8$ & $54.0 \pm 0.8$ \\
SVM+ \cite{Vapnik09} & $75.0 \pm 0.2$ &  $79.4 \pm 0.3$ & $48.5 \pm 0.1$ \\
\textbf{LUPI-HCRF}& $\mathbf{84.9} \pm 0.8$ &  $\mathbf{85.4} \pm 0.4$ & $\mathbf{58.1} \pm 1.4$ \\
\hline
\end{tabular}
}
\end{center}
\caption{Comparison of the classification accuracies ($\%$) on
TVHI, SBU and USAA datasets using hand-crafted features.} \label{Tab:table3}
\end{table}
Table \ref{Tab:table3} compares the proposed approach with
state-of-the-art methods on the human activity classification
task on the same datasets. The results indicate that our
approach improved the classification accuracy. On TVHI, we
significantly managed to increase the classification accuracy
by approximately $10\%$ with respect to the SVM+ approach.
Also, the improvement of our method with respect to SVM+ was
about $6\%$ and $10\%$ for the SBU and USAA datasets,
respectively. This indicates the strength of the LUPI paradigm
and demonstrates the need for additional information.

\textbf{Classification with CNN features:} In our experiments,
we used CNNs for both end-to-end classification and feature
extraction. In both cases, we employed the pre-trained model of
Tran \etal \cite{tran2015learning}, which is a 3D ConvNet. We
selected this model because it was trained on a very large
dataset, namely Sports 1M dataset \cite{Karpathy_2014_CVPR},
which provides very good features for the activity recognition
task, especially in our case where the size of the training
data is small.

For the SBU dataset, which is a fairly small dataset, only a
few parameters had to be trained to avoid overfitting.
We replaced the fully-connected layer of the
pre-trained model with a new fully-connected layer of size
$1024$ and retrained the model by adding a softmax layer on top
of it. For the TVHI dataset, we fine-tuned the last group of
convolutional layers, while for USAA, we fine-tuned
the last two groups. Each group has two convolutional layers,
while we added a new fully-connected layer of size $256$. For
the optimization process, we used mini-batch gradient descent
(SGD) with momentum. The size of the mini-batch was set to $16$
with a constant momentum of $0.9$. For the SBU dataset, the
learning rate was initialized to $0.01$ and it was decayed by a
factor of $0.1$, while the total number of training epochs was
$1000$. For the TVHI and USAA datasets, we used a constant
learning rate of $10^{-4}$ and the total number of training
epochs was $500$ and $250$, respectively. For all datasets, we
added a dropout layer after the new fully-connected layer with
probability $0.5$. Also, we performed data augmentation on each
batch online and $16$ consecutive frames were randomly selected
for each video. These frames were randomly cropped, resulting
in frames of size $112 \times 112$ and then flipped with
probability $0.5$. For classification, we used the centered
$112 \times 112$ crop on the frames of each video sequence.
Then, for each video, we extracted $10$ random clips of $16$
frames and averaged their predictions. Finally, to avoid
overfitting, we used early stopping and extracted CNN features
from the newly added fully-connected layer.

\begin{table}[!t]
\begin{center}
\resizebox{0.99\columnwidth}{!}{
\begin{tabular}{l|ccc}
\hline
\multicolumn{4}{c}{CNN features}\\
\hline
Method & TVHI & SBU & USAA \\
\hline  \hline
CNN (end-to-end) \cite{tran2015learning}  &$60.5 \pm 1.1$& $94.2 \pm 0.8$ & $67.4 \pm 0.6$ \\
SVM \cite{Bishop06}  & $90.0 \pm 0.3$ & $92.8 \pm 0.2$ & $91.9 \pm 0.3$ \\
HCRF \cite{Quattoni07}  & $89.6 \pm 0.5$ & $91.1 \pm 0.4$ & $91.6 \pm 0.8$ \\
SVM+ \cite{Vapnik09}  & $92.5 \pm 0.4$ & $94.8 \pm 0.3$ & $92.3 \pm 0.3$ \\
\textbf{LUPI-HCRF} & $\mathbf{93.2} \pm 0.6$ & $\mathbf{94.9} \pm 0.7$ & $\mathbf{93.9} \pm 0.9$ \\
\hline
\end{tabular}
}
\end{center}
\caption{Comparison of the classification accuracies ($\%$) on
TVHI, SBU and USAA datasets using CNN features.}
\label{Tab:table4}
\end{table}

Table \ref{Tab:table4} summarizes the comparison of LUPI-HCRF
with state-of-the-art methods using the features extracted from
the CNN model, and end-to-end learning of the CNN model using
softmax loss for the classification. The improvement of
accuracy, compared to the classification based
on the traditional features (Table \ref{Tab:table3}), indicates
that CNNs may efficiently extract informative features without
any need to hand design them. We may observe that privileged
information works in favor of the classification task in all
cases. LUPI-HCRF achieves notably higher recognition accuracy
with respect to the HCRF model and the SVM+
approaches. Moreover, for both TVHI and USAA datasets, when
LUPI-HCRF is compared to the end-to-end CNN model, it achieved
an improvement of approximately $33\%$ and $27\%$,
respectively. This huge improvement can be explained by the
fact that the CNN model uses a very simple classifier in the
softmax layer, while LUPI-HCRF is a more sophisticated model
that can efficiently handle sequential data in a more
principled way.

\begin{figure}[!t]
\begin{center}
\begin{tabular}{cc}
    \multicolumn{2}{c}{\includegraphics[width=0.4\columnwidth, clip=true]{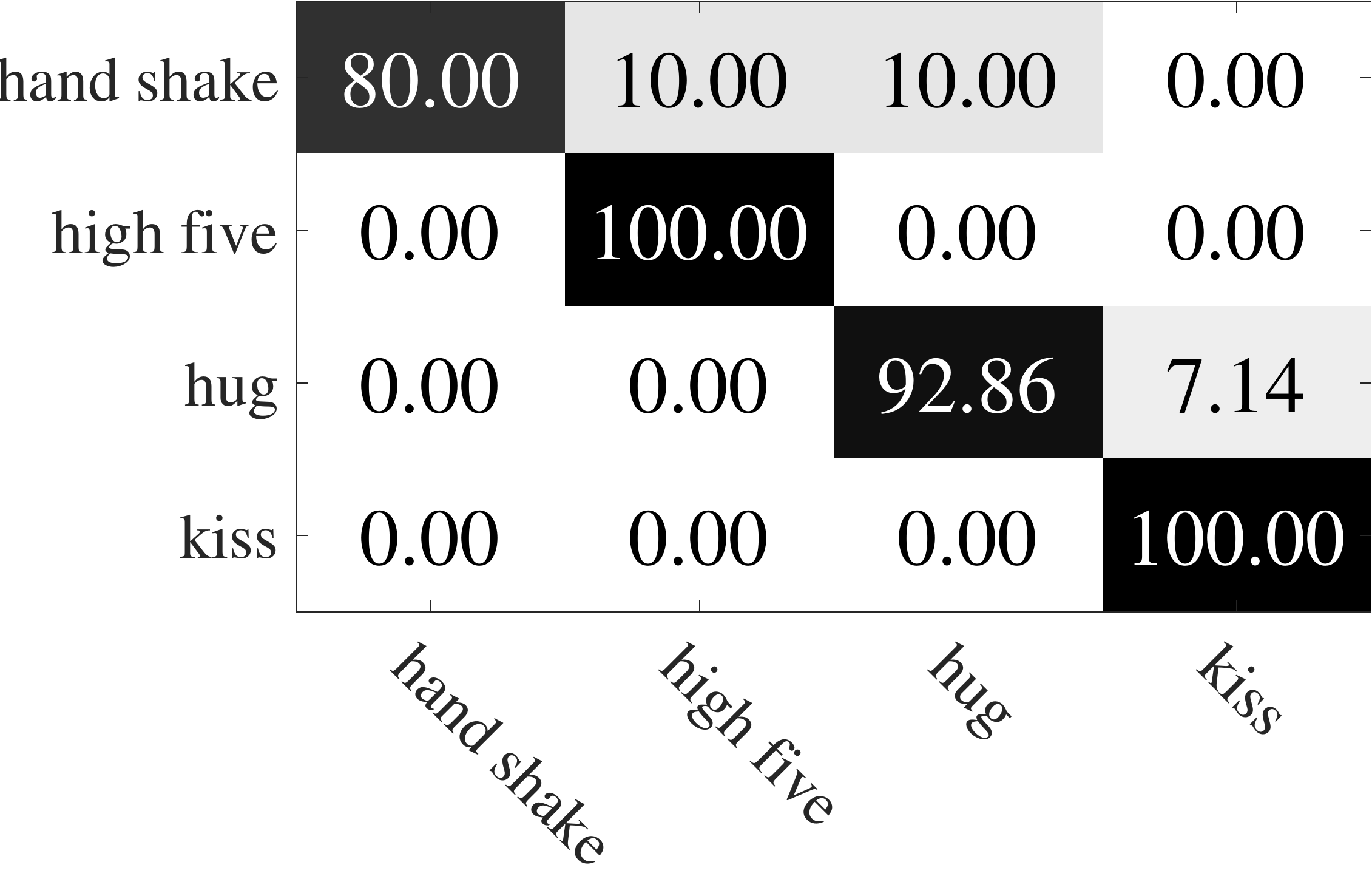} }\\
    \multicolumn{2}{c}{(a) } \\  %~ \\
    \includegraphics[width=0.47\columnwidth, clip=true]{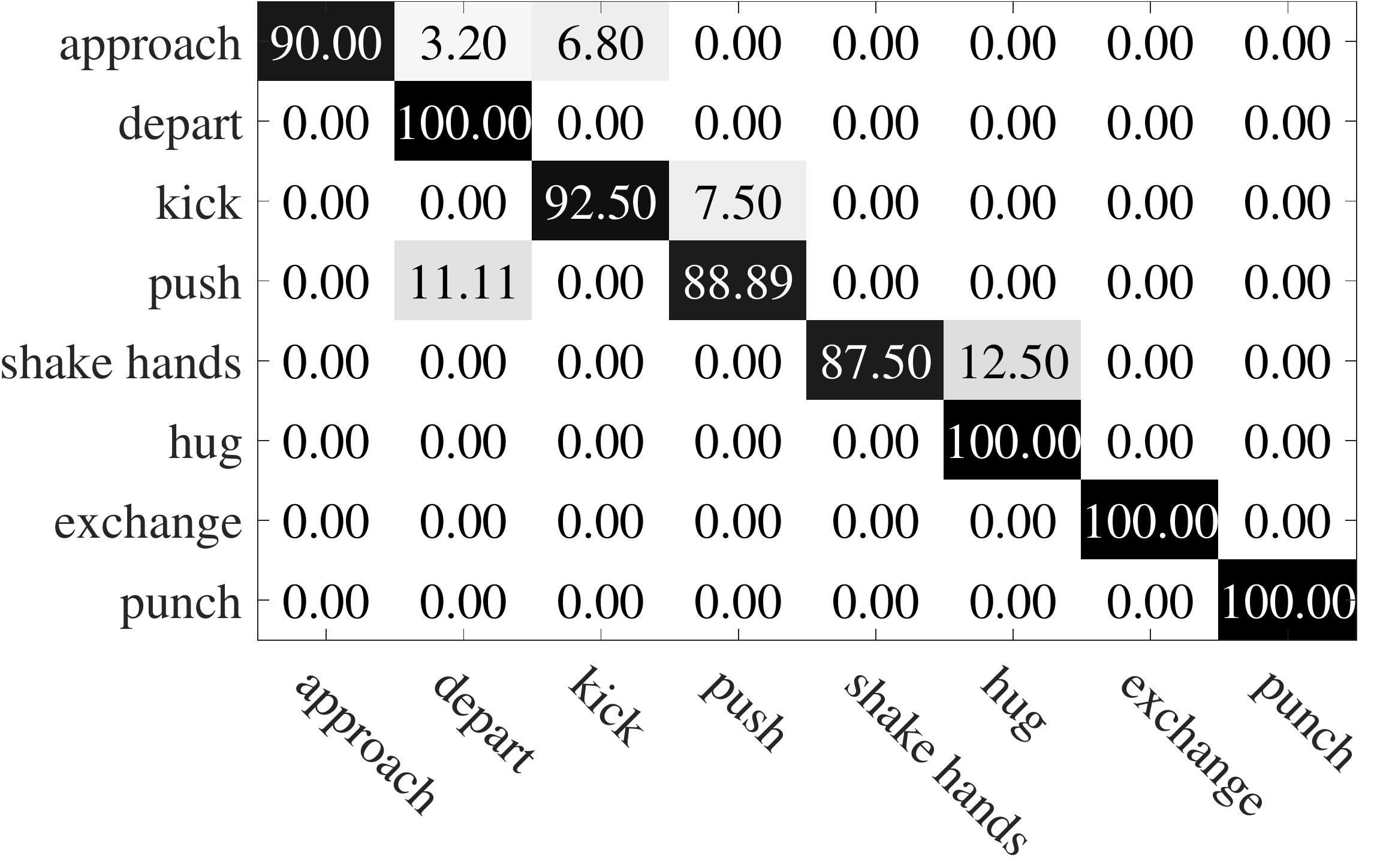} &
    \includegraphics[width=0.47\columnwidth, clip=true]{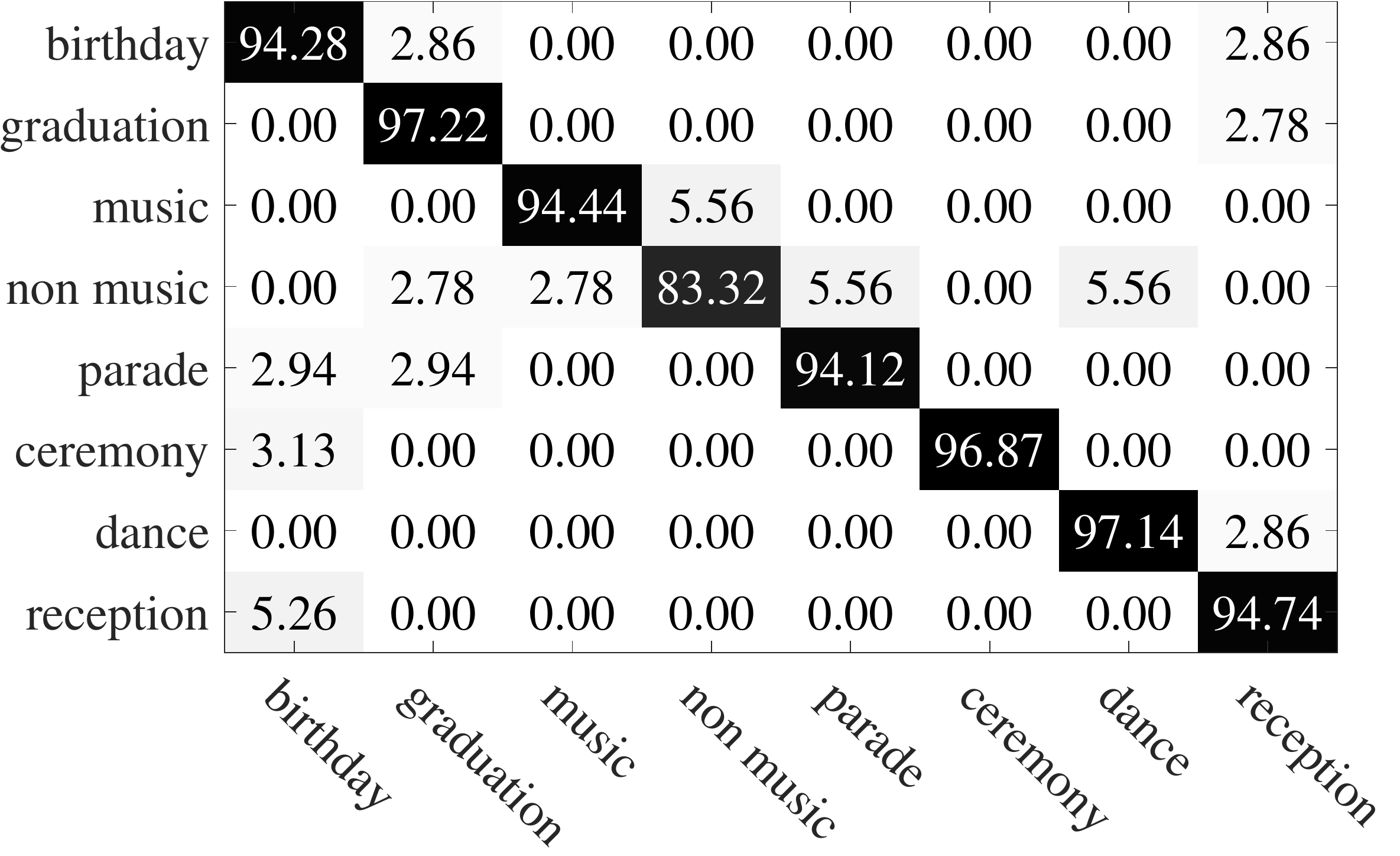} \\
    (b) & (c)
\end{tabular}
\end{center}
    \caption{Confusion matrices for the classification results of the
    proposed LUPI-HCRF approach for (a) the TVHI \cite{Perez12}, (b) the SBU
    \cite{Ykiwon12}, and (c) the USAA \cite{FuHXG12} datasets using the CNN features.}
\label{fig:figure5}
\end{figure}

The corresponding confusion matrices using the CNN-based features, are depicted
in Fig. \ref{fig:figure5}. The combination of privileged
information with the information learned from the CNN model
feature representation resulted in very small inter- and
intra-class classification errors for all datasets.

\textbf{Discussion:} In general, our method is able to robustly
use privileged information in a more efficient way than its
SVM+ counterpart by exploiting the hidden dynamics between the
videos and the privileged information. However, selecting which
features can act as privileged information is not
straightforward. The performance of LUPI-based classifiers
relies on the delicate relationship between the regular and the
privileged information. Tables \ref{Tab:table3} and \ref{Tab:table4}
show that SVM and HCRF perform worse than LUPI-HCRF. This is
because at training time privileged information comes as ground truth but at test
time it is not. Also, privileged information is costly
or difficult to obtain with respect to producing additional
regular training examples \cite{SerraToro14}.

%------------------------------------------------------------------------
\section{Conclusion}
\label{sec:conclutions}

In this paper, we addressed the problem of human activity
recognition and proposed a novel probabilistic classification
model based on robust learning by incorporating a Student's
\textit{t}-distribution into the LUPI paradigm, called
LUPI-HCRF. We evaluated the performance of our method on three
publicly available datasets and tested various forms of data
that can be used as privileged. The experimental results
indicated that robust privileged information ameliorates the
recognition performance. We demonstrated improved results with
respect to the state-of-the-art approaches that may or may not
incorporate privileged information.

\section*{Acknowledgments}
{\small\noindent{This work was funded in part by
the UH Hugh Roy and Lillie Cranz Cullen Endowment Fund and by
the European Commission (H2020-MSCA-IF-2014), under grant
agreement No. 656094. All statements of fact, opinion or
conclusions contained herein are those of the authors and
should not be construed as representing the official views or
policies of the sponsors.}}

{\small
\bibliographystyle{ieee}
\bibliography{ICCV2017_LUPIHCRF_bib}

}

\end{document}